\documentclass{article}
\usepackage{graphicx} 
\usepackage{authblk}
\usepackage{hyperref}
\usepackage{caption}

\usepackage[backend=biber,maxnames=3]{biblatex}
\title{MAIA: A Collaborative Medical AI Platform for Integrated Healthcare Innovation}
\author[1,2]{Simone Bendazzoli}
\author[1]{Sanna Persson}
\author[2,3,6]{Mehdi Astaraki}
\author[4,7]{Sebastian Pettersson}
\author[4,7]{Vitali Grozman}
\author[1,5]{Rodrigo Moreno}
\affil[1]{\small  Department of Biomedical Engineering and Health Systems, KTH, Royal Institute of Technology, Stockholm, Sweden}
\affil[2]{\small Department of Clinical Sciences, Intervention and Engineering, Karolinska Institutet, Stockholm, Sweden}
\affil[3]{\small Division of Medical Radiation Physics, Department of Physics, Stockholm University, Stockholm, Sweden}
\affil[4]{\small Department of Radiology, Karolinska University Hospital, Stockholm, Sweden}
\affil[5]{\small Department of Neurobiology, Care Sciences and Society, Karolinska Institutet, Stockholm, Sweden}
\affil[6]{\small Department of Oncology-Pathology, Karolinska Institutet, Stockholm, Sweden}
\affil[7]{\small Department of Molecular Medicine and Surgery, Karolinska Institutet, Stockholm, Sweden}

\date{}
\begin{document}

\maketitle

\begin{abstract}
The integration of Artificial Intelligence (AI) into clinical workflows requires robust collaborative platforms that are able to bridge the gap between technical innovation and practical healthcare applications. This paper introduces MAIA (Medical Artificial Intelligence Assistant), an open-source platform designed to facilitate interdisciplinary collaboration among clinicians, researchers, and AI developers. Built on Kubernetes, MAIA offers a modular, scalable environment with integrated tools for data management, model development, annotation, deployment, and clinical feedback. Key features include project isolation, CI/CD automation, integration with high-computing infrastructures and in clinical workflows. MAIA supports real-world use cases in medical imaging AI, with deployments in both academic and clinical environments. By promoting collaborations and interoperability, MAIA aims to accelerate the translation of AI research into impactful clinical solutions while promoting reproducibility, transparency, and user-centered design. We showcase the use of MAIA with different projects, both at KTH Royal Institute of Technology and Karolinska University Hospital.
\end{abstract}


\section{Introduction}

Artificial Intelligence (AI) integration in healthcare has emerged as a transformative force, promising to revolutionize patient care, optimize resource allocation, and enhance clinical decision-making \cite{Bajwa2021,MalekiVarnosfaderani2024}. As the healthcare ecosystem increasingly recognizes the importance of AI-powered tools, there is a growing need for collaborative platforms to facilitate the development, deployment, and management of AI solutions in medical settings \cite{hitachi2021haip,oberhuber2017federated}. Modern healthcare institutions are facing complex challenges that demand sophisticated technological solutions. A comprehensive Medical AI Platform can serve as a powerful foundation for addressing these complex needs, effectively bridging technological capabilities with clinical requirements.

One of the open challenges in healthcare is the management of the vast amounts of data handled in clinical settings.
Cloud-based medical AI platforms can provide new opportunities for computational resource sharing, enabling institutions to optimize data storage, and collaborative research environments. By creating a unified and standardised ecosystem, these platforms break down traditional institutional barriers, facilitating knowledge exchange between medical professionals, data scientists, and researchers.
Additionally, the complexity of healthcare AI development necessitates a collaborative approach. 

A shared platform can promote cooperation between medical professionals, data scientists and researchers.
This collaborative environment can accelerate innovation and improve the quality of AI solutions in healthcare \cite{carter2024ai}.

The key advantage of such a platform is its capacity to support the full AI development lifecycle, offering an end-to-end solution. From data collection and preprocessing to model training, validation, deployment, and continuous development, it provides an integrated environment that promotes innovation in medical AI.

A critical objective of these platforms is transforming theoretical research into practical, clinically relevant solutions. By offering structured frameworks for AI model development and deployment, they enable the continuous integration of innovative technologies into existing clinical workflows, while actively involving clinical expertise can enhance compliance with ethical and regulatory standards.

Finally, the success of such medical AI platforms relies on their ability to prioritize end-user experiences. Intuitive interfaces, transparent decision-making processes, and highly customizable solutions are essential to gaining widespread adoption among healthcare professionals \cite{ai_sweden_2024_healthcare,Cai2019}.

\subsection{Existing Solutions}
Despite the need for tools to ease the development, deployment, and collaboration between technical and clinical people for AI-based image analysis solutions,  only a few solutions are available at the moment. We review some of them in this section.

\textbf{OpenShift AI\footnote{https://www.redhat.com/en/products/ai/openshift-ai}:} 
This is a commercial solution from Red Hat. Although OpenShift AI offers powerful AI development and deployment capabilities, it falls short in several key areas.
Firstly, as a proprietary platform, its underlying algorithms and AI-based decision-making processes may not be fully transparent, which might be problematic in healthcare settings where explainability is crucial. Additionally, these platforms often lack specialized tools and frameworks specifically designed for medical AI development, as they are typically designed toward a more generic scientific computing audience. This lack of domain-specific tools can limit the efficiency and effectiveness of medical-focused AI projects. Furthermore, commercial solutions can be prohibitively expensive for research institutions and smaller healthcare providers. Finally, concerns about data sovereignty and privacy may arise when using commercial cloud-based platforms for sensitive medical data. 

\textbf{MONAI\footnote{https://monai.io}:} 
Open-source initiatives like MONAI (Medical Open Network for AI)\cite{monai} offer more collaborative and transparent approaches. These ecosystems allow for greater customization, community-driven development, and the ability to run on-premises, addressing many commercial solutions' limitations. Open-source platforms also promote a more democratic approach to AI development in healthcare, enabling a wider range of institutions to participate in and benefit from advances in medical AI. The only limitation of MONAI is that it does not provide a comprehensive solution with all the tools available in one platform. This issue is targeted in MAIA.

\textbf{Kaapana\footnote{https://www.kaapana.ai/}:} This is an open-source solution developed by the German Cancer Research Center (DKFZ) \cite{kaapana}. Kaapana is an innovative toolkit for AI model development in the medical field. It offers many features, including modularity, customizability, imaging database management, and advanced nnUNet-based tools for segmentation tasks \cite{nnunet}.
Despite its strengths, Kaapana faces several limitations that impact its versatility and applicability in diverse research environments: Firstly, the platform lacks robust project isolation, which constrains its effectiveness in multi-team or multi-study settings. Secondly, Kaapana's reliance on a specific Kubernetes distribution (\textit{microk8s}) restricts its deployment flexibility, making it incompatible with various Kubernetes distributions and multi-cluster management scenarios. Furthermore, the absence of native integration with continuous integration/continuous deployment (CI/CD) pipelines limits the platform's and its tools' efficient updates and maintenance. Kaapana also falls short in providing a unified workspace that combines scientific packages, integrated development environments (IDEs), and medical visualization tools in a shared environment. Another significant drawback is the lack of integration with the MONAI ecosystem, including MONAI Label\footnote{https://monai.io/label.html} for active learning and NVIDIA NVFlare\footnote{https://nvidia.github.io/NVFlare/} for federated learning projects. This can potentially limit researchers' access to cutting-edge tools in the field of medical AI. Additionally, Kaapana does not offer seamless integration with high-performance computing systems, often crucial in high-end model development workflows. Lastly, the platform lacks comprehensive tools for AI model deployment in clinical contexts, potentially limiting the translation of research into practical applications.

\textbf{Escudero-Sanchez et al.}: In  \cite{EscuderoSanchez2023}, the authors propose an AI platform for medical imaging that emphasizes integration with MONAI, imaging databases, and visualization tools. The platform is built on open-source technologies to promote academic reproducibility, reduce costs, and avoid reliance on proprietary software. It features a zero-footprint architecture, eliminating the need for local installations and simplifying compatibility. The interface is optimized for radiological workflows, incorporating tools such as XNAT\footnote{https://www.xnat.org/}, NVIDIA Clara\footnote{https://www.nvidia.com/en-us/clara/}, and 3D Slicer\footnote{https://www.slicer.org/}. It supports multiple AI model types, including fully automated, semi-automated, and interactive models like Deep Grow. The platform is validated through a case study on ovarian cancer segmentation and treatment response prediction. However, its current application is limited to that specific use case, and neither source code nor deployment resources are publicly available.

\subsection{Contribution}

In this paper, we introduce the Medical Artificial Intelligence Assistant (MAIA), a platform designed to promote open collaboration among researchers, clinicians, data scientists, and healthcare professionals. By removing traditional barriers and promoting interdisciplinary knowledge exchange, the objective is to create an inclusive ecosystem where innovative ideas can develop, bridging the gap between theoretical research and practical clinical applications while maintaining high scientific and ethical standards.
The MAIA platform focuses on supporting multiple projects, offering a collaborative environment where AI developers and clinicians can co-create and deploy AI solutions, with a specific focus on integration into existing clinical settings.

The MAIA platform is open source and available on GitHub at \href{https://github.com/kthcloud/MAIA}{\path{github.com/kthcloud/MAIA}}, with comprehensive documentation at \href{https://maia-toolkit.readthedocs.io}{\path{maia-toolkit.readthedocs.io}}, and an online instance accessible at \href{https://maia.app.cloud.cbh.kth.se}{\path{maia.app.cloud.cbh.kth.se}}.

 \newpage
\section{Implementation}

MAIA, the proposed Medical AI platform, aims to establish a collaborative ecosystem in healthcare by integrating expertise from AI technology development and clinical practice. Its primary objectives include promoting knowledge exchange, building long-term collaborations, and advancing research in the medical AI field. 

The core principles of the MAIA platform are:

\begin{itemize}
    \item \textbf{Enhancing AI Education}: Providing learning opportunities for healthcare professionals and students, who are approaching the Medical AI field, with hands-on tutorials and essential development AI tools to build practical skills.
    \item \textbf{Integrating Research}: Bridging AI advancements with medical research to promote innovation.
    \item \textbf{Deploying AI in Clinics}: Implementing AI solutions within clinical workflows to drive meaningful improvements in patient care and operational efficiency through the developed AI solutions.
\end{itemize}

\subsection{Platform Architecture}
\subsubsection{Backbone}
MAIA is designed using Kubernetes as the underlying infrastructure for scalability, security, and efficient resource management. 
The platform supports various Kubernetes distributions (including \textit{MicroK8s\footnote{https://microk8s.io/}, Rancher\footnote{https://www.rancher.com/}, K3S\footnote{https://k3s.io/}}, and \textit{K0S}\footnote{https://k0sproject.io/}) and emphasizes modularity through \textit{Helm Charts}\footnote{https://helm.sh/}, allowing independent installation and deployment of all MAIA modules.

\begin{figure}[!h]
    \centering
    \includegraphics[width=0.7\linewidth]{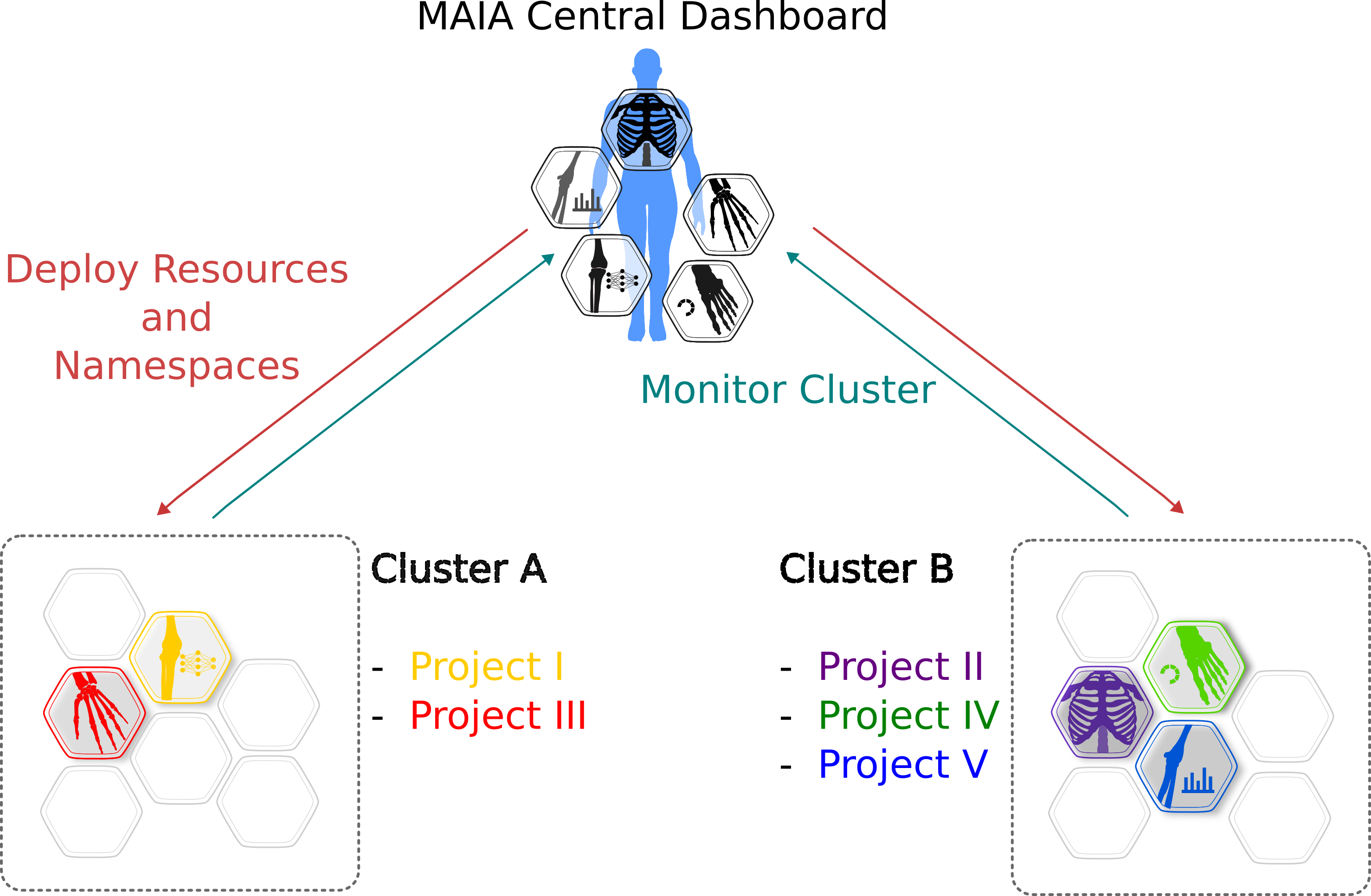}
    \caption{\textbf{Federation of Cluster Representation}: The central MAIA dashboard connects to remote MAIA clusters, allowing users and administrators to view and manage their active projects. Internally, the MAIA Dashboard orchestrates projects by distributing them across available clusters based on individual resource requests (GPUs, memory, and CPU cores). It allocates resources, deploys workloads to remote clusters, and collects monitoring logs to track cluster status and availability.}
    \label{fig:maia_clusters}
\end{figure}

Furthermore, the MAIA platform employs a "Federation of Clusters" architecture (Figure \ref{fig:maia_clusters}), enabling deployment across multiple network-independent infrastructures. This design facilitates inter-connectivity between isolated clusters, promoting cross-infrastructure modularity and flexible resource allocation.

Additionally, MAIA's architecture prioritizes adaptability to different infrastructures, adapting the MAIA modules to the underlying infrastructure. This design enables the adoption of MAIA in on-premises environments, where it can integrate with pre-existing authentication mechanisms for secure user authentication.

Finally, the MAIA architecture integrates ArgoCD\footnote{https://argo-cd.readthedocs.io} to implement CI/CD practices, ensuring the consistent management across various infrastructures. This approach enables automated synchronization of the deployed modules, facilitating reliable and efficient application updates.

\begin{figure}[!h]
    \centering
    \includegraphics[width=0.7\linewidth]{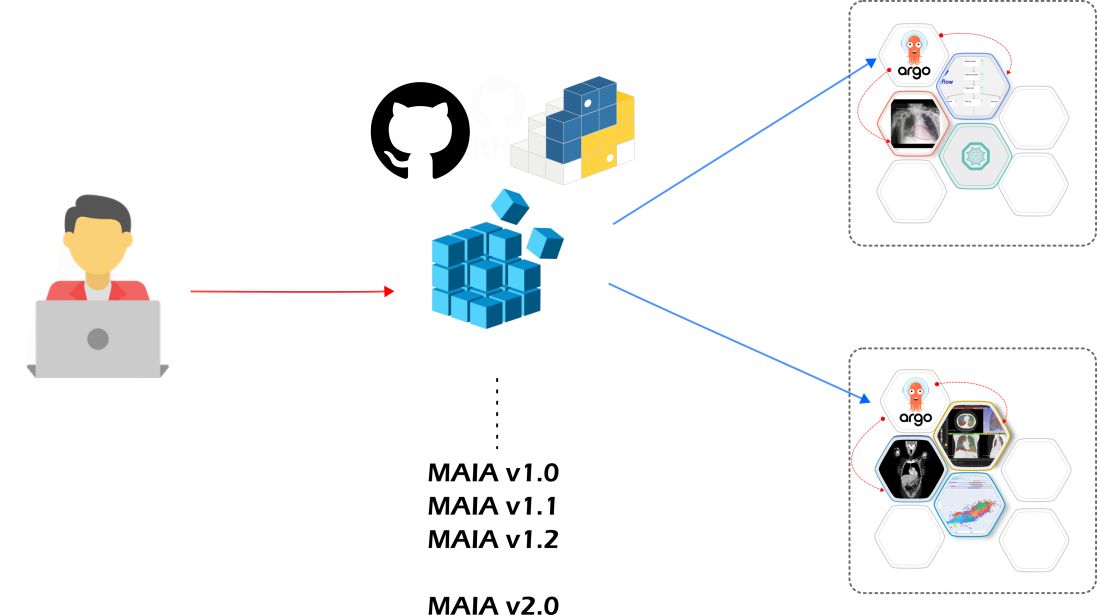}
    \caption{\textbf{MAIA CI/CD}: MAIA developers upgrade, package, and distribute MAIA applications from the central MAIA registry. Individual MAIA instances can then retrieve the updated application versions and apply upgrades either automatically or on a scheduled basis.}
    \label{fig:ci-cd}
\end{figure}

As illustrated in Figure \ref{fig:ci-cd}, MAIA development is packaged in subsequent releases. On the cluster side, each ArgoCD instance continuously monitors these MAIA releases as part of the Continuous Integration workflow, automatically initiating upgrades for individual components as new versions become available.

\subsubsection{MAIA Namespaces}
\begin{figure}[!h]
    \centering
    \includegraphics[width=0.9\linewidth]{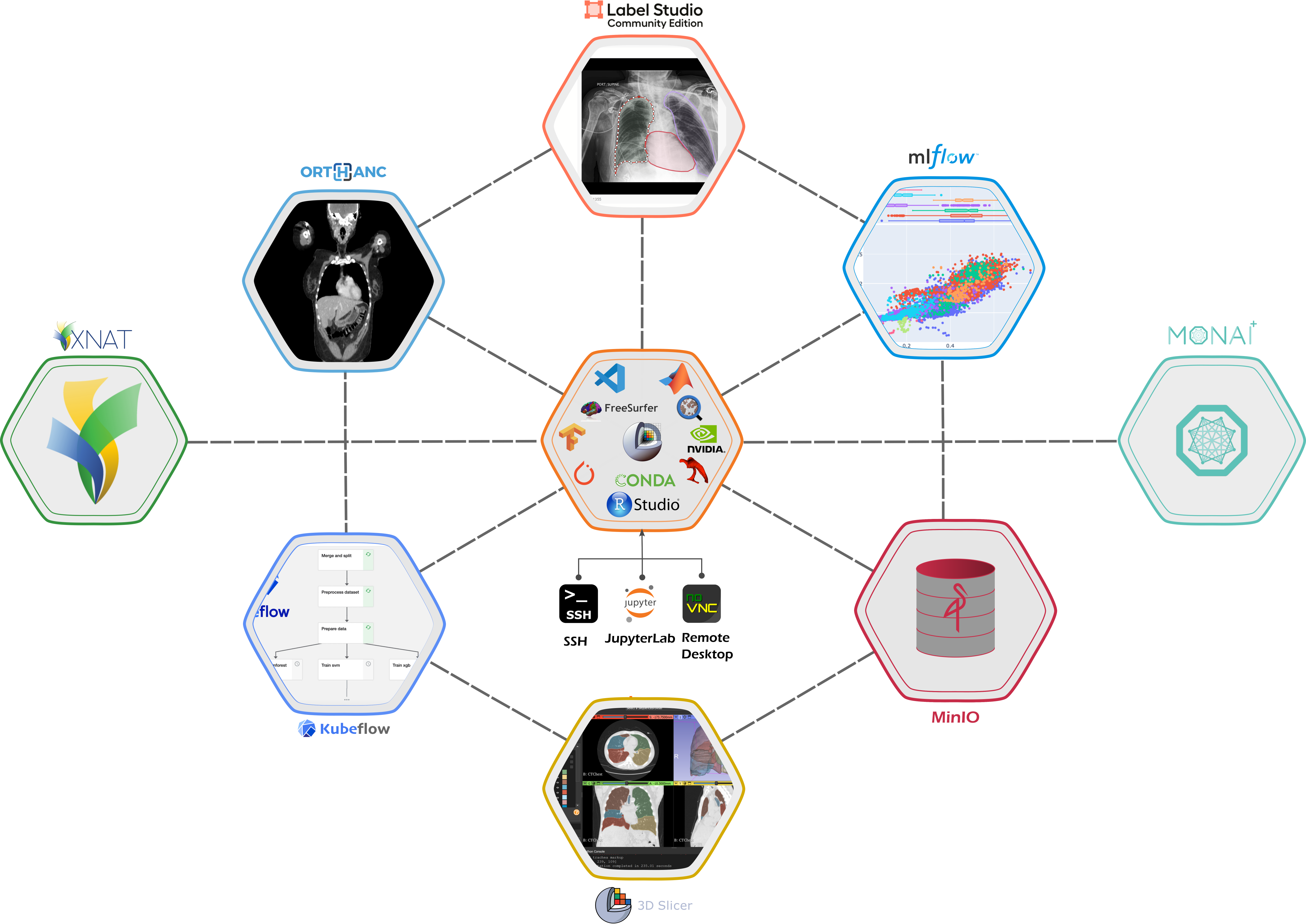}
    \caption{\textbf{MAIA Namespace}: The central MAIA workspace serves as the main entry point for users, with all other applications deployed within the same namespace and integrated.}
    \label{fig:maia-workspace}
\end{figure}
The MAIA platform is structured to accommodate multiple projects, each bringing together users who share a common research objective, promoting collaboration and teamwork among them. The main objective is to share information, data, and resources among researchers and clinicians while maintaining isolated environments for each project. These isolated environments, called \textbf{MAIA Namespaces}, are built upon the Kubernetes concept of namespaces.

MAIA Namespaces serve as virtual entities hosting individual projects. Within each namespace, all project applications are hosted and run, requesting and allocating resources from the underlying Kubernetes namespace infrastructure. This design ensures that each project has access to all necessary tools within its scope while remaining independent and isolated from other project environments.

The MAIA Namespace concept is structured to provide equal authorizations and privileges to all members of a project group. This approach allows for efficient collaboration within projects while maintaining separation between different research initiatives.

\subsubsection{MAIA Namespace Modules}

Each MAIA namespace is provided with a comprehensive suite of integrated tools, including distinct applications for research applications and collaborative workflows.

\paragraph{Core Components}
\begin{itemize}
\item \textbf{MAIA Workspace}: The central hub of the MAIA Namespace, serving as the user's entry point and connecting all other available applications. (Further details are provided in Subsection \ref{workspace}.)
\item \textbf{MLFlow\footnote{https://mlflow.org/}}: A tool for logging and monitoring machine learning and deep learning experiments, ensuring reproducibility. It also functions as a registry for trained models.

\item \textbf{MinIO\footnote{https://min.io/}}: A cloud storage service allowing users to upload, download and share files of any generic format.
\end{itemize}

\paragraph{Medical Imaging Tools}
\begin{itemize}
\item \textbf{Orthanc\footnote{https://www.orthanc-server.com/} and OHIF Viewer\footnote{https://viewer.ohif.org/}}: Orthanc is an open-source DICOM picture archiving and communication system (PACS) for storing and sharing medical images. OHIF Viewer is a web-based DICOM viewer that is integrated with Orthanc.
\end{itemize}

\paragraph{AI and Machine Learning Support}
\begin{itemize}
\item \textbf{KubeFlow\footnote{https://www.kubeflow.org/}}: A platform for building and deploying portable machine learning workflows, including model training, deployment, and pre/post-processing pipelines. This enables easy sharing and reproduction of different workflows.
\item \textbf{XNAT\footnote{https://www.xnat.org/}}: A tool designed to integrate AI-based applications into clinical deployment scenarios.
\item \textbf{MONAI Deploy\footnote{https://monai.io/deploy.html}}: A MONAI-based framework for integrating automatic AI-powered pipelines within existing clinical workflows.
\end{itemize}

\paragraph{Data Annotation}
\begin{itemize}
\item \textbf{Label Studio\footnote{https://labelstud.io/}}: An application for annotating various types of data, including images and audio tracks.
\end{itemize}

\subsubsection{MAIA Workspace}\label{workspace}
The MAIA Workspace is the central component of each MAIA Namespace, offering users a comprehensive environment for developing medical AI applications. Built on JupyterHub\footnote{https://jupyter.org/hub}, it enables each user to launch their own JupyterLab server, providing an isolated workspace with allocated resources specific for AI model development, including GPU-powered environments when requested. The centralized platform facilitates collaborative work in data science and machine learning projects, particularly those focused on medical AI.
The MAIA Workspace offers multiple access points, including a  Jupyter interface, a remote desktop interface, and SSH access for integration with various IDE solutions.

The workspace is equipped with a range of scientific computing environments, such as Visual Studio Code\footnote{https://code.visualstudio.com/}, MATLAB\footnote{https://www.mathworks.com/products/matlab.html}, RStudio\footnote{https://posit.co/products/open-source/rstudio/}, and Anaconda\footnote{https://www.anaconda.com/}.  It also features a curated collection of tutorials designed to guide users through fundamental deep learning examples, optimize the use of the available tools within the MAIA Namespace, and provide comprehensive instruction on the full range of workflows that can be implemented within the MAIA platform.

Based on the single project requirements, optional tools can be automatically installed in the MAIA Workspace. These include specialized software such as QuPath\footnote{https://qupath.github.io/} for histopathology imaging, FreeSurfer\footnote{https://surfer.nmr.mgh.harvard.edu/} for neuroimaging processing, and 3D Slicer\footnote{https://www.slicer.org/} for visualization and processing of medical images.

\subsubsection{Admin Layers}
Alongside MAIA Namespace, MAIA's architecture incudes two internal layers: MAIA Core and MAIA Admin, serving the main purpose of platform management and maintenance.
These layers are dedicated to cluster and platform orchestration, including all the operations related to user and project management, cluster logging and monitoring, cluster spawning, and managing the internal platform networking system.

\begin{figure}[!h]
    \centering
    \includegraphics[width=0.9\linewidth]{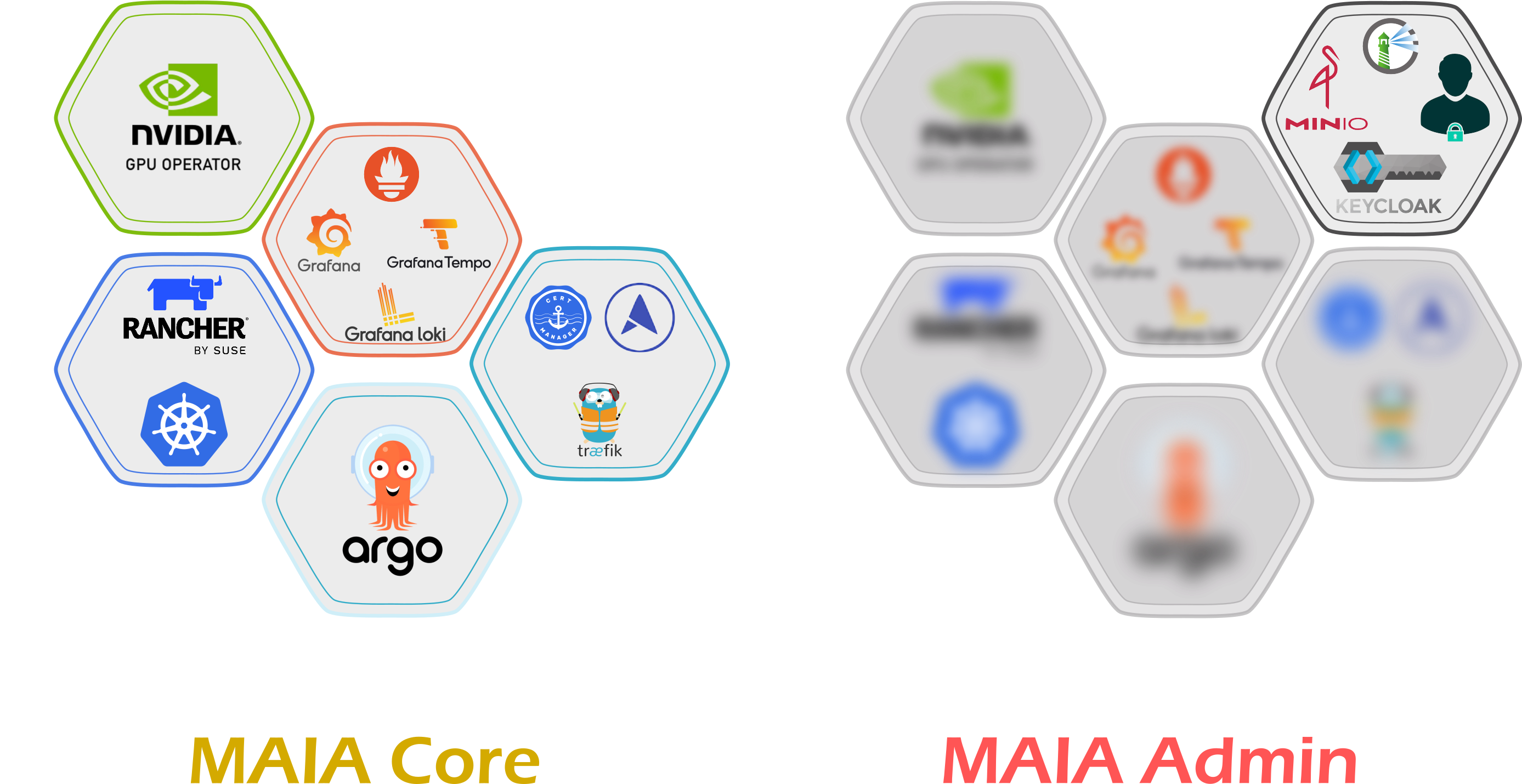}
    \caption{The MAIA Core Layer (left) manages the essential aspects of the Kubernetes platform. On top of it, the MAIA Admin Layer (right) is deployed to configure the Kubernetes cluster for hosting the MAIA infrastructure.}
    \label{fig:core-admin}
\end{figure}

\subsubsection{MAIA Core}\label{maia:core}
MAIA Core is responsible for the foundational aspects of deploying and managing the platform, especially when establishing MAIA in a new environment.
The MAIA Core includes modules for:
\begin{itemize}
\item \textbf{Deployment and Management}
\begin{itemize}
\item \textbf{ArgoCD\footnote{https://argo-cd.readthedocs.io}}: A declarative GitOps continuous delivery tool CI/CD in Kubernetes
\item \textbf{MinIO Operator\footnote{https://min.io/}}: A tool for deployment and management of high-performance object storage in Kubernetes.
\item \textbf{NFS Provisioner\footnote{https://kubernetes-sigs.github.io/nfs-subdir-external-provisioner/}}: A tool to create persistent volumes in pre-configured NFS servers.
\end{itemize}

\item \textbf{Networking and Load Balancing}
\begin{itemize}
\item \textbf{Traefik\footnote{https://traefik.io/traefik/}}: A tool to intercept and manage all the incoming traffic, rerouting it to the cluster network.
\item \textbf{MetalLB\footnote{https://metallb.io/}}: A load-balancer implementation for bare metal Kubernetes clusters using standard routing protocols.
\end{itemize}

\item \textbf{Monitoring and Logging}
\begin{itemize}
\item \textbf{Grafana\footnote{https://grafana.com/}}: A multi-platform open source analytics and interactive visualization web application.
\item \textbf{Loki\footnote{https://grafana.com/oss/loki/}}: A horizontally-scalable, highly-available, multi-tenant log aggregation system.
\item \textbf{Prometheus\footnote{https://prometheus.io/}} A monitoring system and time series database for metrics collection and alerting.
\item \textbf{Tempo\footnote{https://grafana.com/oss/tempo/}}: A distributed tracing backend for observability in microservices environments.
\end{itemize}

\item \textbf{Security and Hardware Support}
\begin{itemize}
\item \textbf{Cert Manager\footnote{https://cert-manager.io/}}; A tool to automate the management and issuance of transport layer security (TLS) certificates in Kubernetes.
\item \textbf{NVIDIA GPU Operator\footnote{https://docs.nvidia.com/datacenter/cloud-native/gpu-operator/}}: A tool for the deployment and management of NVIDIA GPU-enabled applications in Kubernetes.
\end{itemize}
\end{itemize}

\subsubsection{MAIA Admin}\label{maia:admin}
  MAIA Admin focuses on user and project management, authentication, and authorization. It controls the MAIA platform across linked clusters, and monitors resources and traffic. 
  
MAIA Admin includes several key components that provide administrative functionality for the MAIA platform:
\begin{itemize}
    \item \textbf{MAIA Dashboard}: A web-based interface that enables users to register projects, request resources, and access various MAIA services deployed on the Kubernetes cluster. It serves as a central hub for interacting with the MAIA platform.
    \item \textbf{Harbor\footnote{https://goharbor.io/}}: An open-source container registry that allows users to securely store, manage, and distribute container images.
    \item \textbf{Keycloak\footnote{https://www.keycloak.org/}}: An open-source identity and access management tool that manages users and roles associated with the MAIA Projects. It provides robust authentication and authorization capabilities.
    \item \textbf{Rancher\footnote{https://www.rancher.com/}}: A Kubernetes management platform that simplifies the deployment, management, and scaling of containerized applications. It offers a user-friendly interface for cluster management and application deployment across various environments.
\end{itemize}

\subsection{Integrating MAIA into Existing Clusters }

As introduced in \ref{maia:core} and \ref{maia:admin}, MAIA's modular design enables adaptability across various existing solutions. Specifically, the \textbf{MAIA Core} module serves as the foundational component for initializing a cluster in a new environment, equipping it with the essential tools required for MAIA's functionality. Meanwhile, the \textbf{MAIA Admin} layer is designed to facilitate cross-cluster operations by synchronizing and unifying tasks such as user management. Additionally, it prepares the MAIA interface to deploy end-user MAIA namespaces across multiple clusters.

To integrate MAIA into an existing cluster, the administrator can deploy the required components from \textbf{MAIA Core} and update the configuration within the existing \textbf{MAIA Admin} layer to register the new cluster accordingly.

\subsection{Integrating MAIA into HPC Systems}

When working with complex deep learning models and high volumes of 3D medical imaging data, researchers often face limitations with their in-house storage and computational resources, such as GPUs, RAM, and CPU cores, which may be insufficient for training some of the state-of-the-art models. In such cases, high-performance computing (HPC) systems offer a viable alternative to overcome these constraints.

In Sweden, researchers can access the National Academic Infrastructure for Supercomputing in Sweden (NAISS)\footnote{https://www.naiss.se/} through the  Swedish User Portal for Research (SUPR) platform. NAISS is a national research infrastructure that provides large-scale high-performance computing resources, storage capacity, and advanced user support for Swedish research.

Individual researchers and research teams can request specific computational allocations for defined periods via the SUPR portal. This process allows for a targeted distribution of resources based on the needs of the project.

HPC systems are designed for maximum efficiency in resource allocation. Projects receive a finite amount of computation time per project, and resources are typically allocated through non-interactive job submission rather than real-time usage.
Once approved, researchers can remotely access the HPC system, transfer necessary data and code, and submit jobs to perform tasks such as model training.

A significant limitation in utilizing HPC systems is the specialized knowledge required. Many systems use queue-based frameworks like SLURM\footnote{https://slurm.schedmd.com/}, and effective use demands proper training for all users. Understanding job submission, resource requests, and queue management is crucial. This learning curve can impact the efficient use of allocated resources, highlighting the importance of user education in HPC environments.

MAIA-HPC is a dedicated submodule of MAIA designed to integrate MAIA with any generic HPC system. It simplifies the processes of transferring data, trained models, and code, as well as submitting jobs. By utilizing MAIA-HPC, users can interact directly with MAIA while the backend handles all implementation details related to data transfer, job submission, and monitoring. This allows MAIA users to collect, prepare, and preprocess data locally on MAIA, transfer it to the HPC system for running experiments, and then retrieve the trained models for validation and result analysis on their local MAIA setup.

MAIA-HPC has been tested by interfacing with multiple HPC systems within NAISS. These systems include Dardel\footnote{https://www.pdc.kth.se/hpc-services/computing-systems/dardel-hpc-system/dardel-1.1043529}, Berzelius\footnote{https://www.nsc.liu.se/systems/berzelius/}, and Alvis\footnote{https://www.naiss.se/resource/alvis/}, showcasing MAIA-HPC's ability to manage diverse workloads across different HPC environments.

Furthermore, MAIA-HPC has extended its capabilities to Bianca\footnote{https://www.naiss.se/resource/bianca/}, a specialized HPC system designed for processing sensitive data. This expansion highlights MAIA-HPC's adaptability in handling not only high-performance computing tasks but also workloads involving sensitive information, making it a versatile tool for researchers across various domains.
\begin{figure}[!h]
    \centering
    \includegraphics[width=0.9\linewidth]{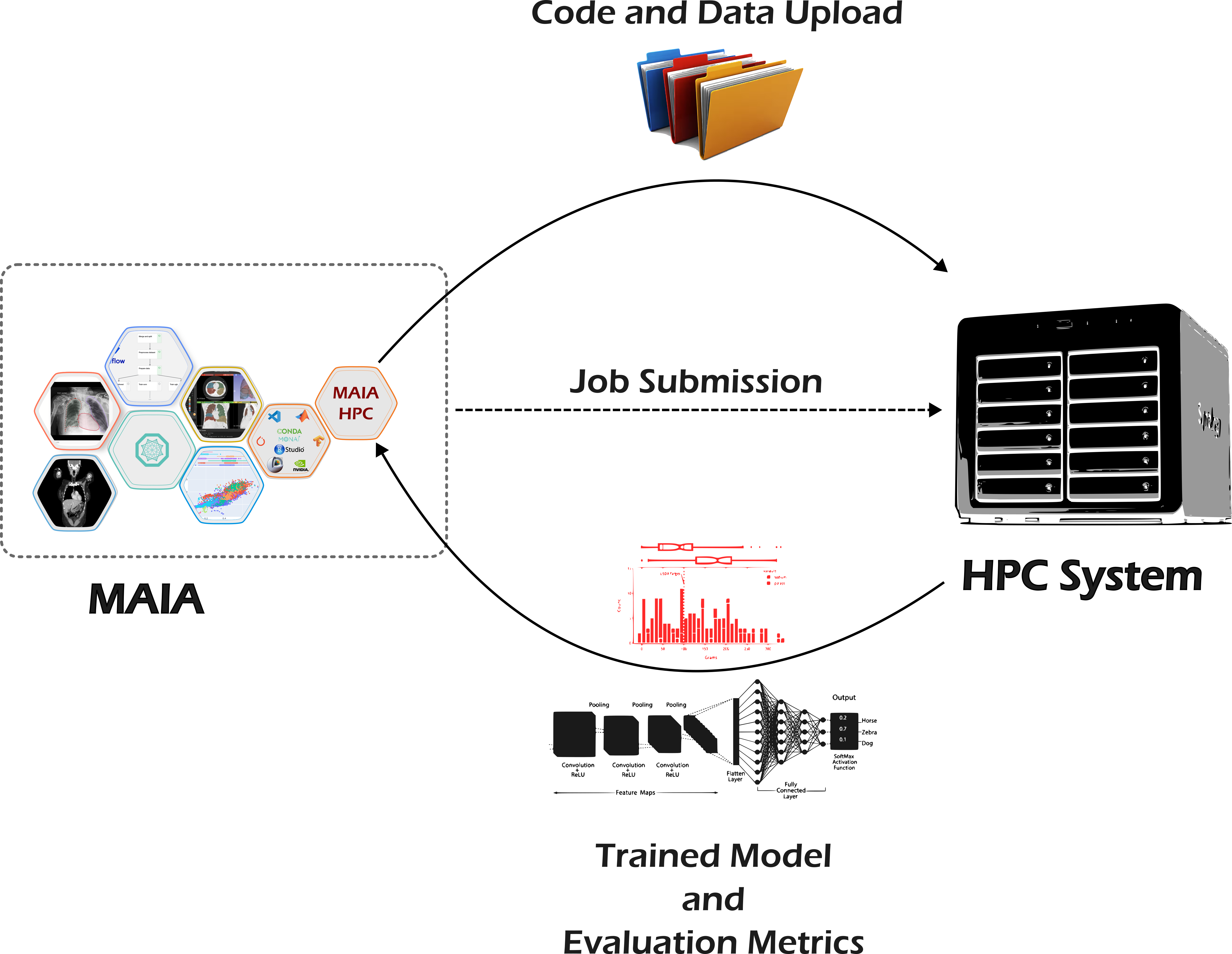}
    \caption{MAIA-HPC Integration: The MAIA-HPC installable module enables smooth connection between MAIA and an HPC system. It streamlines the process of uploading code and data in a structured manner and provides a simple command interface for job submission. The module also monitors job status and, upon successful completion, allows users to retrieve the results of the job execution.}
    \label{fig:maia-hpc}
\end{figure}
\subsubsection{GPU booking system}\label{gpu_booking_section}
One of the main use cases for MAIA is as a platform for continuous development and testing of machine learning models. On a project level, the development is often done in sprints followed by less intense use of GPU resources. Recognizing this structure, the utilization of a MAIA cluster can be highly variable, with some GPUs being idle during periods of low research activity, even though the project still requires workspace availability and resource continuity.

\begin{figure}[!ht]
    \centering
    \includegraphics[width=0.9\linewidth]{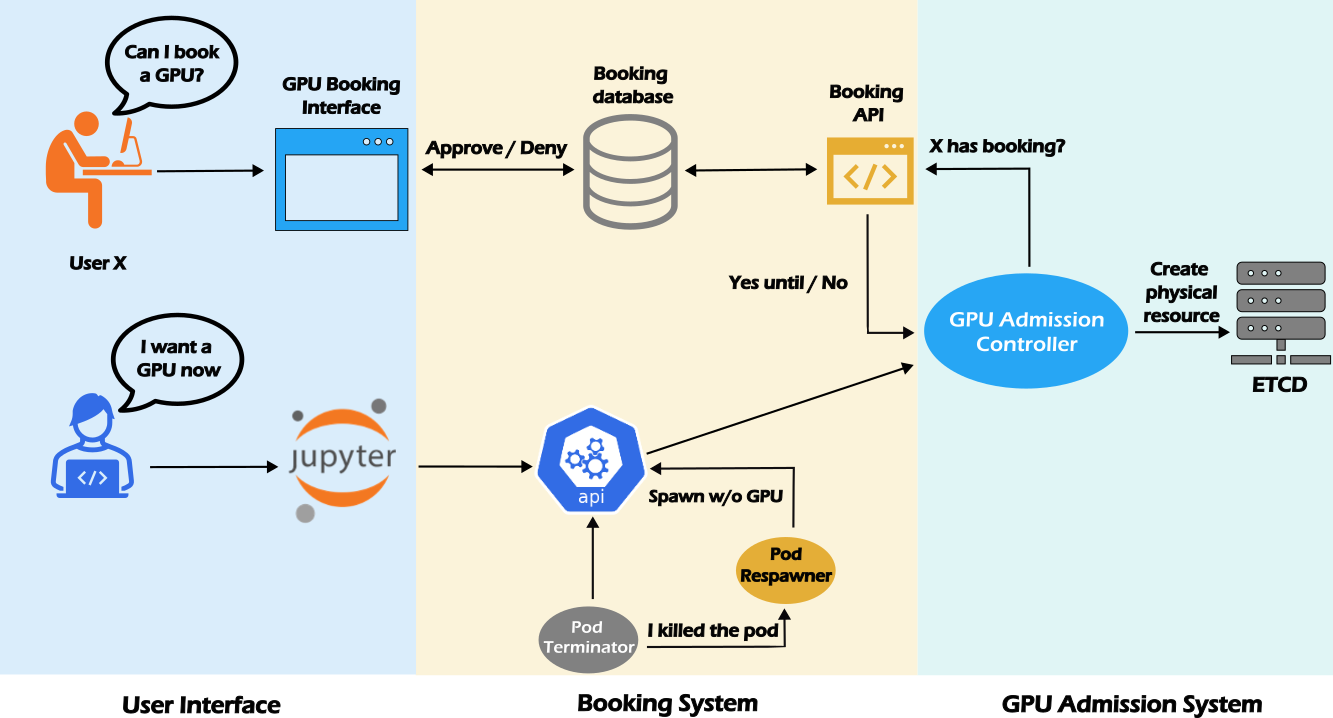}
    \caption{Users book GPU sessions through the booking interface and start them via Jupyter. The system verifies bookings via the booking API and GPU admission controller, provisioning GPUs through Kubernetes if valid. After expiration, pods are terminated and respawned without GPU access.}
    \label{fig:booking}
\end{figure}

To address this variability and increase resource utilization, we implemented a GPU booking system. Users can reserve GPUs for specific periods through a dedicated booking interface (see Figure \ref{fig:booking}). When users initiate their booked sessions via Jupyter, the system communicates with the booking API and the GPU admission controller to verify the reservation and allocate GPU resources accordingly through Kubernetes. At the end of a booking period, the user's pod is automatically terminated and respawned without GPU access. This ensures GPUs are freed and returned to the shared pool, available for immediate reallocation to other projects and researchers.

This approach significantly improves the overall utilization of GPU, promotes equitable resource distribution among users, and helps maintain an organized environment that aligns with the dynamic needs of ongoing research and development activities.

\section{MAIA Workflows}

MAIA's modular architecture allows AI researchers and clinicians to define and adapt workflows for various scenarios and applications. Since these modules operate independently, users can create customized workflows without leaving the platform, fully adopting MAIA’s ecosystem to cover all stages of the AI lifecycle. In the following section, we showcase various applications where MAIA has been adopted, detailing the underlying workflows supported by its components.

\subsection{AI Development Workflow}
This workflow centers on the MAIA Workspace module, which provides researchers with AI development tools accessible via three entry points: \textit{SSH}, \textit{Jupyter}, and \textit{Remote Desktop}. Data for model development is transferred into MAIA through \textbf{MinIO}, offering multiple options for server data exchange. \textbf{Kubeflow Pipelines} prepare and move this data into the workspace for integration into the AI routines.
During model training, the workspace connects to \textbf{MLFlow} for monitoring and visualizing evaluation metrics. For image inspection and prediction analysis, researchers can use the remote desktop interface of \textbf{3D Slicer}.
\begin{figure}[!h]
    \centering
    \includegraphics[width=0.9\linewidth]{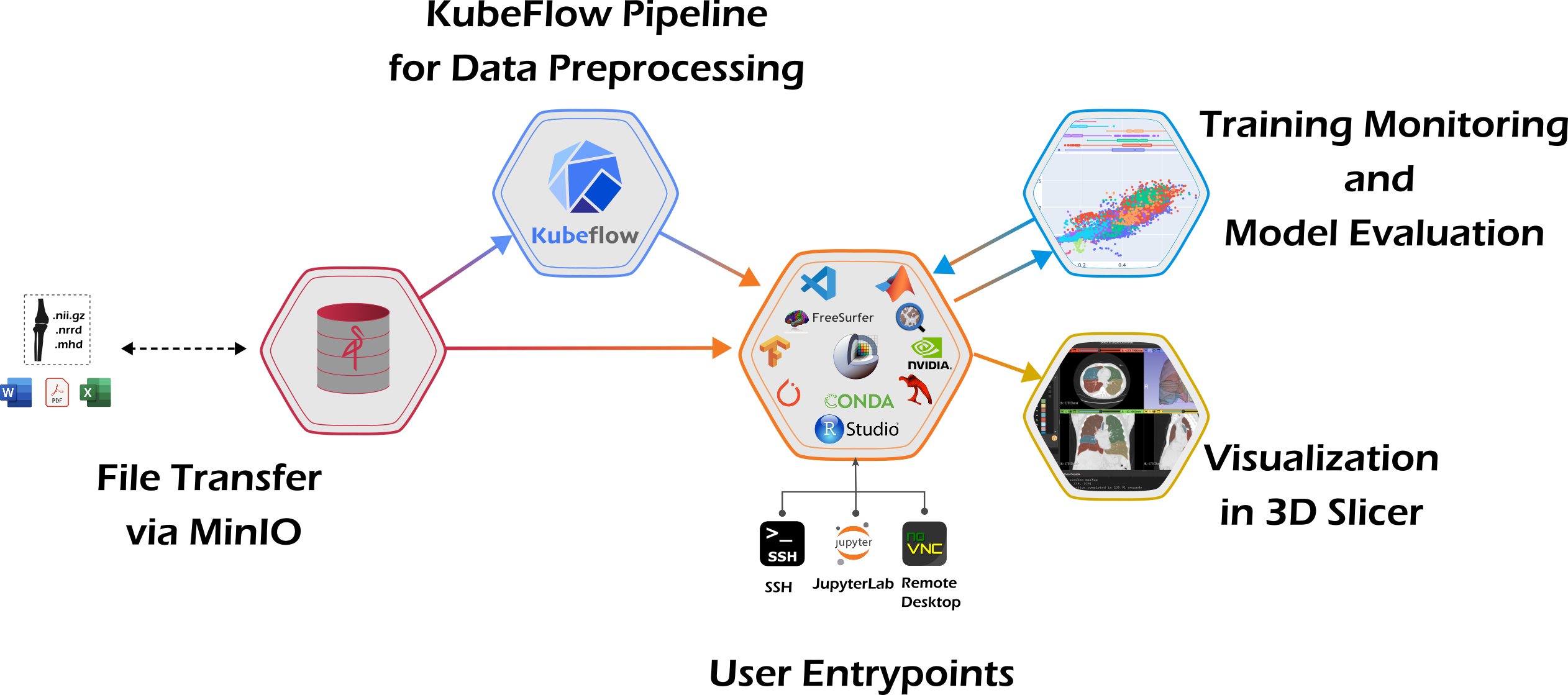}
    \caption{AI Development Workflow: In this workflow, data and files are transferred via \textbf{MinIO}, which is directly accessible from the MAIA workspace. If additional data preparation or preprocessing is required, users can adopt \textbf{KubeFlow} to integrate custom workflows, with the preprocessed data becoming available in the workspace upon completion. During model training, tools such as \textbf{MlFlow} and \textbf{Remote Desktop} support real-time monitoring of training progress, visualization of validation performance, and inspection of model predictions.}
    \label{fig:ai-workflow}
\end{figure}

\subsection{Clinical Environment Workflow}
MAIA has, as mentioned, the ability to comprehensively harbour multiple project functions, including medical imaging databases and AI development workflow tools, combined with project isolation functionality to enable work on many different parallel projects.  Realizing large-scale medical AI development projects in the MAIA namespace requires an easy yet secure method of collecting and exporting large medical image datasets to the MAIA namespace. 

To address this need, the AI group within the IT Department at Karolinska University Hospital (KUH) has created RADIANCE, an image preparation workflow that can search, filter, and semi-automatically export medical DICOM images from the clinical image servers. Such data is then imported into the Orthanc DICOM server within an MAIA Namespace. 

RADIANCE can search all KUH databases for patient records and medical images to identify suitable patient cohorts according to defined criteria, from which clinical and imaging data can be exported with automated pseudonymization.  This export workflow enables easy handling of large image databases while adhering to strict security regarding the personal information of the subjects. The workflow to create a patient cohort and automatically export medical images into MAIA is done according to the following steps:

1. Identification of a suitable patient cohort by searching in RADIANCE the patient records according to an international classification of diagnoses (ICD) code or a combination of ICD codes. 

2. Cross-matching by RADIANCE against defined radiological examination codes to find subjects with appropriate medical imaging available for the project in question. 

3. Filtering of available imaging data according to parameters such as the study time period, age, and sex. 

4. Defining detailed criteria in RADIANCE for DICOM image export in order to exclude unwanted image stacks from each patient/examination. The criteria include image stack plane orientation, slice thickness, minimum images in the stack, and exclusion of free-text search terms in the series description.

5. Selecting the appropriate MAIA-Orthanc server for the project in question to which RADIANCE should transfer the images.

6. The IT Department at KUH performs quality control and checks ethical permits, after which the imaging data DICOM information is automatically pseudonymized and the images transferred to the selected MAIA-Orthanc server. 

Once the DICOM images are transferred into the desired MAIA-Orthanc instance, they can then be visualized and processed with any of the available tools in the MAIA namespace, including XNAT, OHIF Viewer, and the 3D Slicer Remote Desktop interface. All three of these options support the MONAI Label connection, meaning that the images transferred into Orthanc can be directly used for active learning, for example.
\begin{figure}
    \centering
    \includegraphics[width=0.9\linewidth]{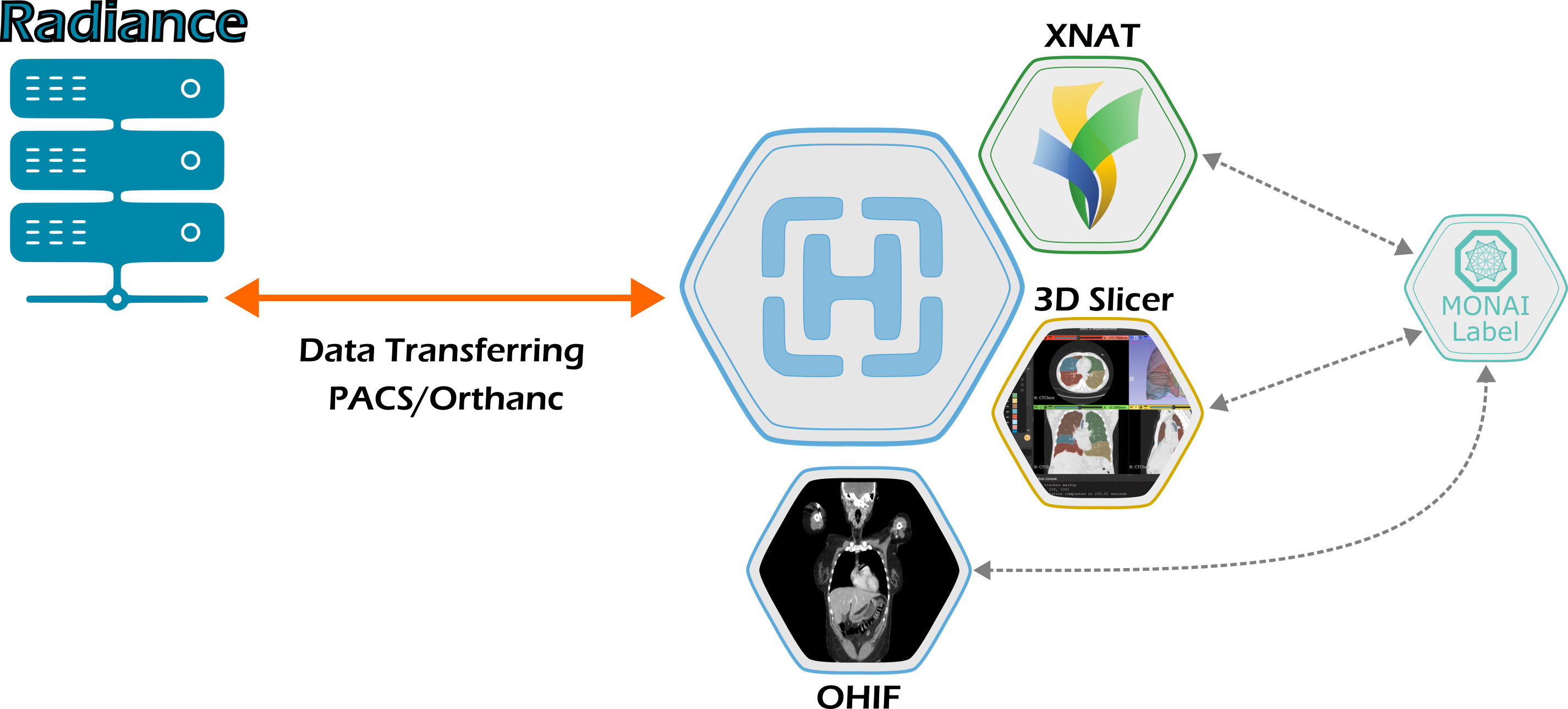}
    \caption{Clinical Environment Workflow: In this workflow, an external PACS—typically hosted within the hospital infrastructure—is connected to the Orthanc instance running in the MAIA namespace, enabling a direct DICOM data flow. Once the DICOM files are received by Orthanc, they become accessible (and optionally processable) through user interfaces such as OHIF, 3D Slicer, and XNAT. These interfaces also provide the ability to initiate Active Learning workflows directly.}
    \label{fig:orthanc}
\end{figure}

\subsubsection{Active Learning Workflow}

In this workflow, the central role is played by radiologists, who are actively validating AI-generated predictions and refining annotations through an iterative feedback process known as active learning (AL). Within the MAIA framework, this is implemented with the \textbf{MONAI Label} component, which integrates AI model training with clinician input.

In the workflow, an AI developer first trains a medical imaging model and deploys it via the \textbf{MONAI Label} tool. Radiologists access the AI predictions through compatible interfaces such as the OHIF Viewer, XNAT, or 3D Slicer. These tools allow direct visualization and modification of AI-generated annotations.

After validation or edits, the updated annotations are sent back to the \textbf{MONAI Label} server, where the model is retrained using this newly labeled data. Over iterations, the system learns from radiologists’ corrections, progressively reducing annotation effort while improving accuracy.
\begin{figure}[!h]
    \centering
    \includegraphics[width=0.9\linewidth]{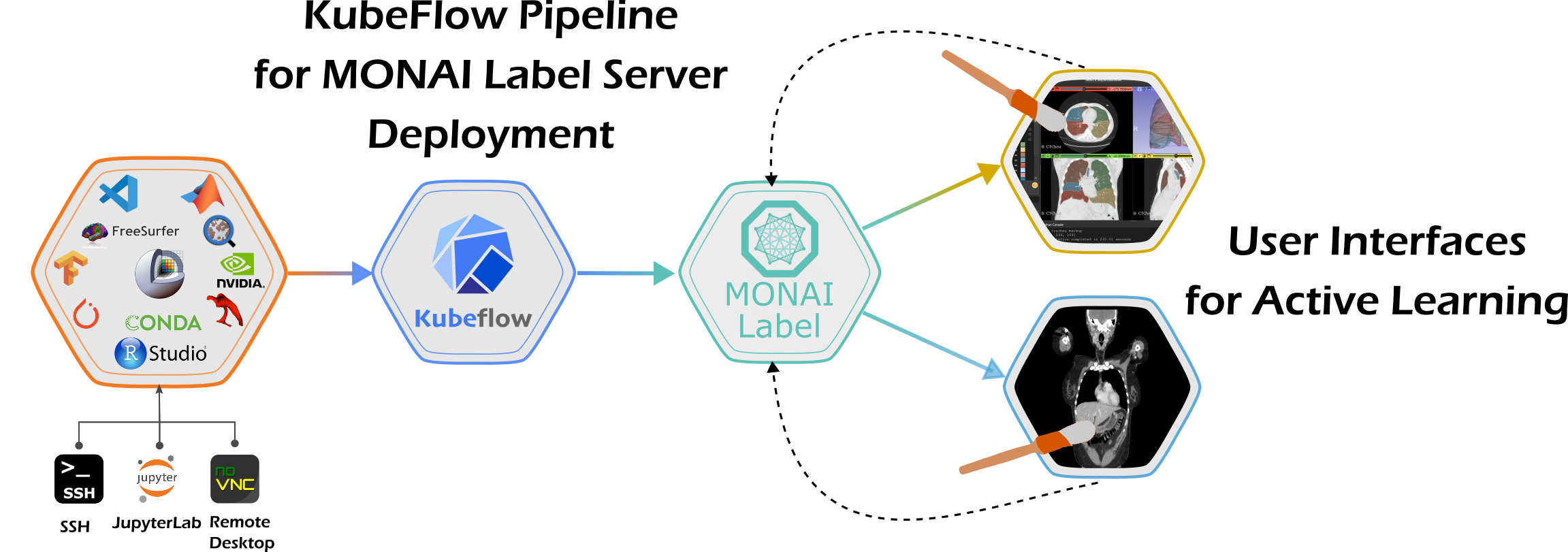}
    \caption{Active Learning Workflow: The workflow begins by launching the \textbf{MONAI Label} server through a \textbf{Kubeflow} pipeline, specifying the model to use and the dataset location for active learning. Once the server is running, users can perform manual annotations or refine model predictions using any of the \textbf{MONAI Label}-compatible interfaces, such as 3D Slicer, OHIF, or XNAT. The updated annotations are sent back to the server, enabling the model to be retrained with the newly labeled data.}
    \label{fig:active-learning}
\end{figure}

One of the key steps in model development from the clinical viewpoint is the testing of the model performance in an environment that is similar to the clinical workflow and the possibility to provide feedback in the same environment. 
The proposed workflow has been successfully validated through the evaluation and re-training of the bone metastasis model, as detailed in Section \ref{bone-mets}. Future iterations will incorporate an active learning component, where the submission of corrected or approved predictions will automatically trigger model re-training. As a result, each new examination reviewed by the radiologist will be analyzed using an increasingly refined model. This simulation of clinical human–AI interaction is important, as it allows radiologists to experience how a model can be integrated into daily practice, supports the evaluation of interactions between multiple models and the clinical workflow, and accelerates the process of model evaluation and re-training while promoting closer collaboration between clinicians and engineers.



\section{Implementation at KTH Royal Institute of Technology}
The MAIA platform is currently implemented at KTH Royal Institute of Technology to support both research and educational computational needs. Specifically, it addresses the requirements of the Department of Biomedical Engineering and Health Systems, serving more than 20 researchers with GPU resources from the cluster federation. Currently, the most powerful GPUs available are NVIDIA RTX A6000 cards, utilized jointly with allocations on national supercomputer clusters, allowing projects to flexibly scale their computational capacity.

Flexibility is further enhanced by integrating a GPU booking system (see Section \ref{gpu_booking_section}), facilitating resource distribution and streamlined onboarding of additional users.

The transition from researchers using individual workstations to a cloud-based system has allowed a standardization of tools, more needs-based division of computational resources and higher utilization of the machines.

\subsection{MAIA in Graduate Research and Education}
The MAIA system is not only a cornerstone for research but also plays a crucial role in graduate education. It has been instrumental in supporting a course in deep learning for biomedical engineering students\footnote{https://www.kth.se/student/kurser/kurs/CM2003?l=en}. Each student group is allocated a GPU and a dedicated workspace for their exercises and course projects. This setup allows for the provision of data, including non-public datasets, in a controlled environment. Pre-installed Python packages and comprehensive technical support are provided, ensuring that students can focus on learning and experimentation without the overhead of managing computational resources. The students are provided with a JupyterHub workspace where they are able to access all of the MAIA tools and use the Visual Studio Code editor directly in their browser. The system also supports users to set up an SSH connection if that is preferred. Administrators of the course have direct access to all students' workspaces, simplifying lab assistance. The system's user-friendly interface and robust support have also made it an ideal platform for hosting a general computer science introductory workshop\footnote{https://github.com/kthcloud/Computer-Science-Workshop}, highlighting its versatility and ease of use.

Furthermore, during the past three years, master students writing their thesis and research interns supervised by our researchers have utilized the MAIA system for their primary computational needs in AI. This approach ensures that students are offered the same computational tools as experienced researchers, promoting an equitable and conducive learning environment. For master project students, in particular, MAIA allows for close collaboration between the supervisor and the student, where the supervisor can have direct access to the same workspace as the student for collaboration and assistance. The setup time for a new student has also decreased radically compared to providing a project student with their own machine.

\section{Implementation at Karolinska University Hospital}

KUH has recently established an AI research environment. MAIA has been deployed in such an environment, which has enabled the development and deployment of computational algorithms designed to address clinically relevant problems utilizing real-world clinical data.
Functionally, MAIA serves as an intermediary, facilitating collaboration between researchers and AI developers on one hand, and clinicians on the other, by providing a unified collaborative platform for the integration of their respective expertise.
Consequently, MAIA has become a central catalyst in the process of AI model development, evaluation, validation, and clinical deployment in this AI research environment at KUH. The following sections will present two current applications in which MAIA served as the core infrastructure for the complete design, development, and deployment of AI models tailored to specific medical imaging tasks within a clinical environment.

\subsection{Detection and Segmentation of Vertebra Metastasis in CT Images}\label{bone-mets}

Bone is a prevalent site of metastasis across various malignancies, known as the third most common, with an annual incidence of 18.8 cases per 100,000 individuals and a survival prognosis ranging from months to several years \cite{BoneMetsReviewe}.
While PET-CT offers enhanced visualization of bone metastases, its elevated cost and limited availability necessitate the utilization of alternative imaging modalities.
MRI provides superior sensitivity for detecting lesions within the marrow and surrounding soft tissue structures without exposing patients to ionizing radiation.
Conversely, CT demonstrates heightened sensitivity in detecting alterations in bone morphology, possesses superior spatial resolution, and, critically, serves as the primary imaging modality for cancer screening across numerous cancer entities.
Consequently, the early detection of malignant bone lesions in CT images is paramount for optimizing treatment strategies, improving patient prognosis, and enhancing therapeutic outcomes. However, the manual analysis of thin-slice, high-resolution CT scans for metastatic lesion detection constitutes a labor-intensive and time-consuming process.
Furthermore, metastatic lesions can exhibit significant morphological similarity to healthy bone structures and present as tiny lesions occupying only a few voxels, thereby compounding the challenges associated with manual detection.
Therefore, the development of robust segmentation models of vertebral metastases is of critical importance and has been the motivation of this study.
The training of accurate deep learning segmentation models, however, requires large-scale, meticulously annotated datasets.
To address this challenge, MAIA has facilitated a collaborative project by providing a comprehensive workflow encompassing data management, preprocessing, annotation, model development, and deployment phases. The following steps delineate the implementation of MAIA within this project:

\begin{itemize}
    \item \textbf{Step 1}: The project begins with only 30 labeled subjects. A preliminary segmentation model was developed based on the initial dataset.
    \item \textbf{Step 2}: The preliminary model was integrated into MONAI Label, enabling radiologists to load new batches of unlabeled data, visualize model predictions, and perform corrective annotations.
    \item \textbf{Repeating steps 1 and 2}: These steps were iteratively repeated until the annotated dataset reached 150 cases.
    \item \textbf{Step 3}: The expanded annotated dataset facilitated extensive experimentation to determine a robust segmentation pipeline. This resulted in a substantial improvement in segmentation model sensitivity, increasing from 55\% with 30 cases to 82\% with 150 cases.
    \item \textbf{Step 4}: The components of the optimized segmentation pipeline were deployed using Kubeflow, managing the entire workflow from DICOM loading and NIfTI conversion through preprocessing, segmentation, optional correction, and postprocessing, to the final DICOM output.
    \item \textbf{Step 5}: The workflow was containerized, enabling on-demand instantiation for the prediction on new patient data.
\end{itemize}

\begin{figure}[!ht]
\centering
\includegraphics[width=1\textwidth]{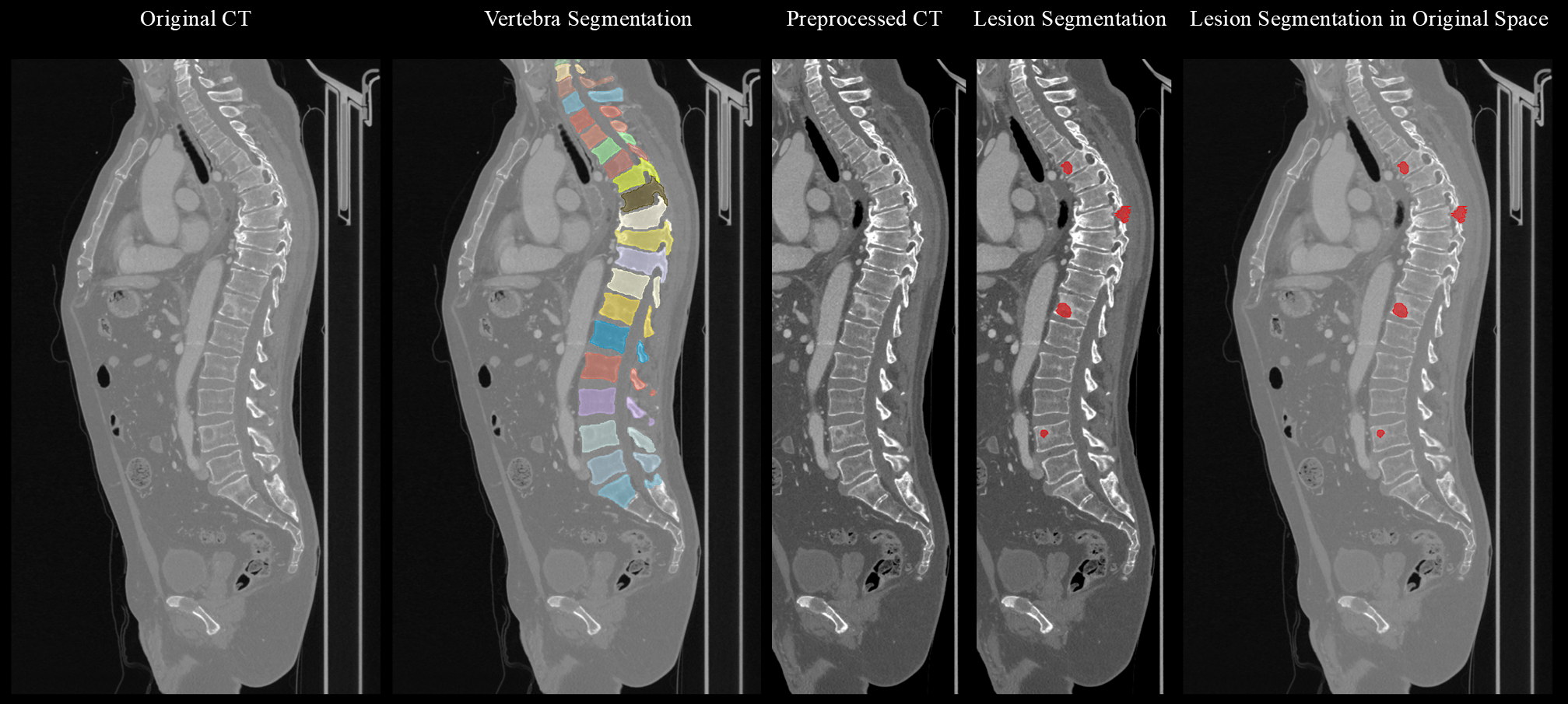}
\caption{The original CT volumes were preprocessed by being cropped around the vertebra region and CT windowing to enhance the contrast prior to segmentation. The final masks were projected into the original CT spacing.}
\label{fig:bone_mets_application}
\end{figure}

Figure \ref{fig:ai-workflow} illustrates the core components of the described workflow, while Figure \ref{fig:bone_mets_application} presents the visual results of the workflow steps (initial vertebra segmentation, followed by bone lesion segmentation on the cropped vertebra region.).

\subsection{Development of an AI Model for Brain Metastasis Segmentation in MRIs}
\label{sec:brainmet}

Brain metastases (BMs) constitute the most prevalent malignancy affecting the adult central nervous system, with an estimated incidence of 20–40\% in patients with systemic cancer \cite{moawad2024braintumorsegmentationbratsmets}.
Given the frequent presentation of multiple lesions at varying stages of treatment, radiologic evaluation often necessitates a longitudinal assessment beyond a single comparative analysis. 
In clinical practice, a comprehensive evaluation commonly involves the review of serial imaging studies to monitor the temporal evolution of metastatic lesions. 
This process, however, can be labor-intensive and time-consuming. Consequently, the development of automated models for the accurate detection, segmentation, of BMs is critical for the formulation of effective therapeutic strategies and accurate prognostication \cite{MLreviewMets}. 
The inherent complexities of BMs, including the heterogeneity in lesion size, shape, and anatomical location, necessitate the implementation of robust methodological approaches capable of identifying such heterogenous metastatic lesions. 
MRI is the standard imaging modality for brain tumor screening. The present study aimed to develop a robust automatic model for the segmentation of BM lesions in MRIs, with the intended application of facilitating prognostic and survival analyses.

To achieve this objective, a segmentation model was developed utilizing datasets provided by the Brain Tumor Segmentation (BraTS) challenge \cite{menze2014multimodal}, with model training performed on the MAIA platform at Karolinska University Hospital.

Specifically, the BraTS-METs 2025 task, focused on BM segmentation, provided a training dataset of 1296 labeled cases. These cases consisted of multi-parametric MRIs (mpMRIs), including non-enhanced T1, post-gadolinium-contrast T1, T2-weighted, and non-enhanced T2-FLAIR volumes, for the segmentation of lesion subregions, encompassing enhancing tissues, non-enhancing tumor core, and surrounding non-enhancing FLAIR hyperintensities, as well as resection cavities.

As with previous applications, the MAIA platform, deployed on-site at Karolinska University Hospital, was used to support the deployment of the pre-trained model.
A key distinction from the vertebral metastasis application lies in the strict preprocessing requirements necessary to align clinical data with the developed model's inputs.
The BraTS dataset underwent extensive preprocessing, necessitating the replication of these steps to achieve the highest performance of the model.
These preprocessing procedures included: conversion of DICOM files to the NIfTI file format, co-registration to the SRI24 template, resampling to a uniform isotropic resolution, and skull stripping.

The implementation of MAIA within this project, adopting the available KubeFlow pipeline module, was executed through the following sequential steps:
\begin{itemize}
    \item \textbf{Step 1}: DICOM file loading and conversion to NIfTI format, adhering to a specific filename convention to maintain the integrity of multi-channel input data. 
    \item \textbf{Step 2}: application of the aforementioned comprehensive preprocessing pipeline.
    \item \textbf{Step 3}: execution of the segmentation pipeline via MONAI Label for the prediction of metastatic lesion segmentation masks and potential post-processing corrections.
\end{itemize}

A schematic representation of the described workflow is presented in Figure \ref{fig:active-learning}. The training of the model was performed using MAIA inside the KUH using its server (an HP ProLiant DL380, with 64 CPU cores and 388 Gb RAM, 2 x Tesla V100S).
\begin{figure}[!ht]
\centering
\includegraphics[width=1\textwidth]{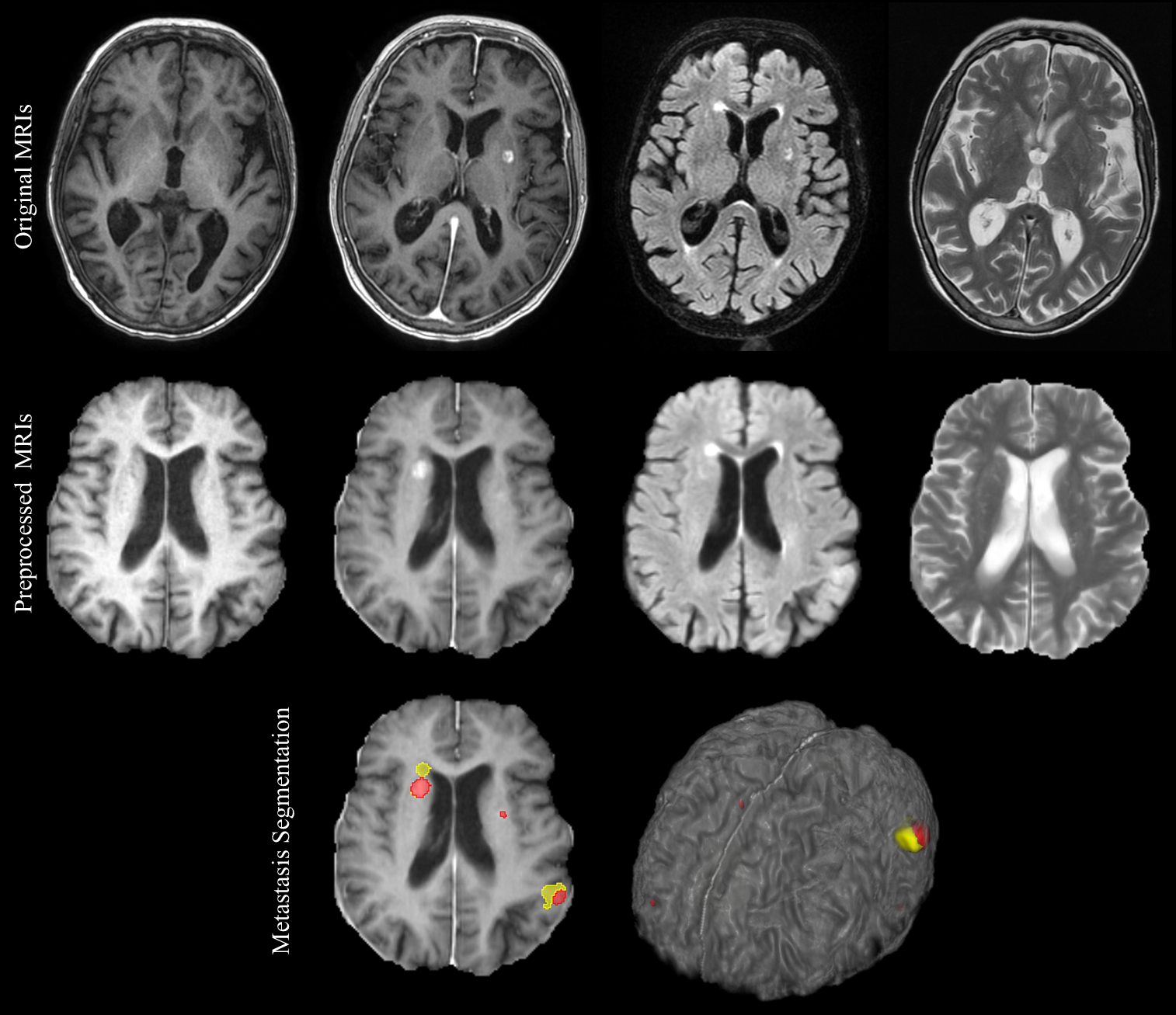}
\caption{Original MRI volumes, preprocessed prior to segmentation. The order of images in the first and second rows are non-enhanced T1w, contrast-enhanced T1w, T2-FLAIR and T2w, respectively.}
\label{fig:brain_mets_application}
\end{figure}
\subsection{AI Deployment and Integration in a Clinical Workflow}

Following the model development and validation phases, the final step involves integrating the model into a clinical workflow—its intended application. The process begins with a clinical PACS sending DICOM data to the MAIA-hosted Orthanc instance, which in turn initiates the MONAI Deploy pipeline. The incoming DICOM image is then analyzed by the trained AI model, packaged as a MONAI Bundle, to identify the target region. The resulting predictions are encapsulated in a DICOM SEG file, which is transmitted back to the clinical PACS, ensuring accessibility for review and use by other downstream clinical tools and systems. This process was performed for the brain metastases project discussed in Sect. \ref{sec:brainmet}. Since the vertebral metastases project from Sect. \ref{bone-mets} includes active learning; it was already integrated into a clinical workflow from the beginning. 

\begin{figure}[!h]
    \centering
    \includegraphics[width=0.9\linewidth]{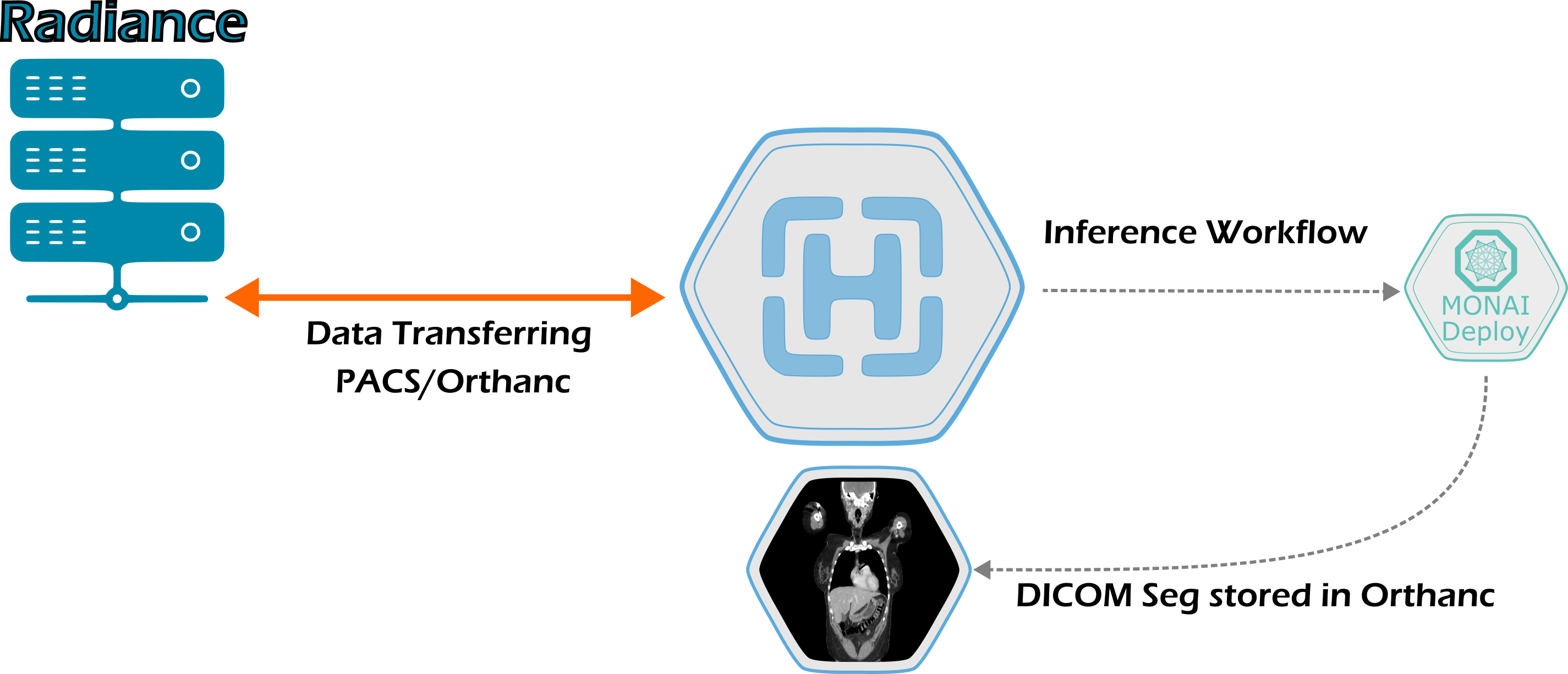}
    \caption{Integration into clinical workflows involves linking the clinical PACS with Orthanc within MAIA. When a DICOM image is transmitted, it automatically triggers the MAIA workflow to process the incoming data. By integrating MONAI Deploy into the pipeline, the system generates an AI prediction, which is then sent back to Orthanc as a DICOM SEG modality.}
    \label{fig:enter-label}
\end{figure}

\section{Impact}
In this paper, we present MAIA, a Medical AI platform designed to promote collaboration in the field and bring AI tools closer to clinical integration within existing workflows. MAIA achieves this by providing a standardized suite of tools and hosting different projects where multidisciplinary expertise can converge and collaborate.

MAIA’s core mission is to bridge the gap between successful AI research and its real-world clinical application. It promotes collaboration by placing clinicians and radiologists at the center of the AI lifecycle, supported by AI scientists. Ultimately, MAIA aims to deliver all the necessary tools in compliance with healthcare security standards, ensuring safe handling of sensitive data. The platform is open-source and designed for on-premise deployment within individual institutions or hospitals, aligning with their specific authentication, authorization, and security protocols.

However, the path to wider adoption and recognition as a standard platform comes with challenges. Sustaining open-source, community-driven interest—or attracting support from stakeholders such as institutions and healthcare providers is essential for ongoing development. Moreover, current clinical testing is limited, and MAIA must further demonstrate its scalability and adaptability across diverse settings and workflows to validate its practical impact in real-world healthcare environments.

To unlock its full potential, MAIA invites the medical AI community, healthcare institutions, and open-source contributors to collaborate in advancing AI-driven clinical care. Through active contribution, testing, and real-world deployment, MAIA aims to accelerate the integration of AI into everyday practice—turning collaborative research into meaningful patient outcomes.
%
\section*{Ethical Approvals}
The components of this study involving bone and brain metastasis, as examined within the MAIA platform, were approved by the Etikprövningsmyndigheten (Swedish Ethical Review Authority) under application number 2022-01893-01. Subsequent amendments were approved under amendment numbers 2023-03637-02, 2024-03784-02, and 2024-08111-02. All procedures were conducted in accordance with applicable ethical guidelines and regulations.

\section*{Funding}  
This study has been partially funded by the Swedish Childhood Cancer Foundation (Barncancerfonden MT2022-0008), by Vinnova through AIDA, project ID: 2319, by Digital Futures, by the Swedish Research Council (Vetenskapsrådet, grant 2022-03389), MedTechLabs, Marie Skłodowska-Curie Doctoral Networks Actions (HORIZON-MSCA-2021-DN-01-01; 101073222), Cancerfonden (22-2389 Pj), HKUST-KTH Global Knowledge Network Awards, Hälsa, Medicin och Teknik (grant No. 2022-0688) and Hjärt-Lungfonden (grant No. 2022-0492). The funders of the study had no role in the study design nor the collection, analysis, and interpretation of data, writing of the report, or decision to submit the manuscript for publication.
 
\section*{Acknowledgements}

We thank the National Academic Infrastructure for Supercomputing in Sweden (NAISS) and the Knut and Alice Wallenberg Foundation for the computational resources at Alvis, Dardel, and Berzelius supercomputers.

\printbibliography

@misc{hitachi2021haip,
  author       = {Hitachi, Ltd. and Nihon Unisys, Ltd. and IBM Japan Ltd. and SoftBank Corp. and Mitsui \& Co., Ltd.},
  title        = {Launch of "Healthcare AI Platform Collaborative Innovation Partnership (HAIP)" Approved by MHLW and METI Ministers},
  year         = {2021},
  month        = {April},
  url          = {https://www.hitachi.com/New/cnews/month/2021/04/210401c.pdf},
  note         = {Press release}
}

@online{oberhuber2017federated,
  author    = {Marion Oberhuber},
  title     = {Federated Learning in Healthcare: The Future of Collaborative Clinical and Biomedical Research},
  year      = {2017},
  month     = feb,
  url       = {https://www.owkin.com/blogs-case-studies/federated-learning-in-healthcare-the-future-of-collaborative-clinical-and-biomedical-research},
  note      = {Owkin Blog}
}

@online{carter2024ai,
  author       = {Stacy Carter and Farah Magrabi and Yves Saint James Aquino},
  title        = {Some clinicians are using AI to write health records. What do you need to know?},
  year         = {2024},
  month        = sep,
  url          = {https://theconversation.com/some-clinicians-are-using-ai-to-write-health-records-what-do-you-need-to-know-237762},
  note         = {The Conversation}
}

@online{ai_sweden_2024_healthcare,
  author    = {{AI Sweden}},
  title     = {AI in healthcare gets a boost from Canada-Sweden knowledge exchange},
  year      = {2024},
  month     = jun,
  url       = {https://www.ai.se/en/news/ai-healthcare-gets-boost-canada-sweden-knowledge-exchange},
  note      = {Accessed: 2025-04-11}
}

@article{Cai2019,
  title = {“Hello AI”: Uncovering the Onboarding Needs of Medical Practitioners for Human-AI Collaborative Decision-Making},
  volume = {3},
  ISSN = {2573-0142},
  url = {http://dx.doi.org/10.1145/3359206},
  DOI = {10.1145/3359206},
  number = {CSCW},
  journal = {Proceedings of the ACM on Human-Computer Interaction},
  publisher = {Association for Computing Machinery (ACM)},
  author = {Cai,  Carrie J. and Winter,  Samantha and Steiner,  David and Wilcox,  Lauren and Terry,  Michael},
  year = {2019},
  month = nov,
  pages = {1–24}
}

@article{Bajwa2021,
  title = {Artificial intelligence in healthcare: transforming the practice of medicine},
  volume = {8},
  ISSN = {2514-6645},
  url = {http://dx.doi.org/10.7861/fhj.2021-0095},
  DOI = {10.7861/fhj.2021-0095},
  number = {2},
  journal = {Future Healthcare Journal},
  publisher = {Elsevier BV},
  author = {Bajwa,  Junaid and Munir,  Usman and Nori,  Aditya and Williams,  Bryan},
  year = {2021},
  month = jul,
  pages = {e188–e194}
}

@article{nnunet,
  title = {nnU-Net: a self-configuring method for deep learning-based biomedical image segmentation},
  volume = {18},
  ISSN = {1548-7105},
  url = {http://dx.doi.org/10.1038/s41592-020-01008-z},
  DOI = {10.1038/s41592-020-01008-z},
  number = {2},
  journal = {Nature Methods},
  publisher = {Springer Science and Business Media LLC},
  author = {Isensee,  Fabian and Jaeger,  Paul F. and Kohl,  Simon A. A. and Petersen,  Jens and Maier-Hein,  Klaus H.},
  year = {2020},
  month = dec,
  pages = {203–211}
}

@article{EscuderoSanchez2023,
  title = {Integrating Artificial Intelligence Tools in the Clinical Research Setting: The Ovarian Cancer Use Case},
  volume = {13},
  ISSN = {2075-4418},
  url = {http://dx.doi.org/10.3390/diagnostics13172813},
  DOI = {10.3390/diagnostics13172813},
  number = {17},
  journal = {Diagnostics},
  publisher = {MDPI AG},
  author = {Escudero Sanchez,  Lorena and Buddenkotte,  Thomas and Al Sa’d,  Mohammad and McCague,  Cathal and Darcy,  James and Rundo,  Leonardo and Samoshkin,  Alex and Graves,  Martin J. and Hollamby,  Victoria and Browne,  Paul and Crispin-Ortuzar,  Mireia and Woitek,  Ramona and Sala,  Evis and Sch\"{o}nlieb,  Carola-Bibiane and Doran,  Simon J. and \"{O}ktem,  Ozan},
  year = {2023},
  month = aug,
  pages = {2813}
}

@article{kaapana,
  title = {Joint Imaging Platform for Federated Clinical Data Analytics},
  ISSN = {2473-4276},
  url = {http://dx.doi.org/10.1200/cci.20.00045},
  DOI = {10.1200/cci.20.00045},
  number = {4},
  journal = {JCO Clinical Cancer Informatics},
  publisher = {American Society of Clinical Oncology (ASCO)},
  author = {Scherer,  Jonas and Nolden,  Marco and Kleesiek,  Jens and Metzger,  Jasmin and Kades,  Klaus and Schneider,  Verena and Bach,  Michael and Sedlaczek,  Oliver and Bucher,  Andreas M. and Vogl,  Thomas J. and Gr\"{u}nwald,  Frank and K\"{u}hn,  Jens-Peter and Hoffmann,  Ralf-Thorsten and Kotzerke,  J\"{o}rg and Bethge,  Oliver and Schimm\"{o}ller,  Lars and Antoch,  Gerald and M\"{u}ller,  Hans-Wilhelm and Daul,  Andreas and Nikolaou,  Konstantin and la Fougère,  Christian and Kunz,  Wolfgang G. and Ingrisch,  Michael and Schachtner,  Balthasar and Ricke,  Jens and Bartenstein,  Peter and Nensa,  Felix and Radbruch,  Alexander and Umutlu,  Lale and Forsting,  Michael and Seifert,  Robert and Herrmann,  Ken and Mayer,  Philipp and Kauczor,  Hans-Ulrich and Penzkofer,  Tobias and Hamm,  Bernd and Brenner,  Winfried and Kloeckner,  Roman and D\"{u}ber,  Christoph and Schreckenberger,  Mathias and Braren,  Rickmer and Kaissis,  Georgios and Makowski,  Marcus and Eiber,  Matthias and Gafita,  Andrei and Trager,  Rupert and Weber,  Wolfgang A. and Neubauer,  Jakob and Reisert,  Marco and Bock,  Michael and Bamberg,  Fabian and Hennig,  J\"{u}rgen and Meyer,  Philipp Tobias and Ruf,  Juri and Haberkorn,  Uwe and Schoenberg,  Stefan O. and Kuder,  Tristan and Neher,  Peter and Floca,  Ralf and Schlemmer,  Heinz-Peter and Maier-Hein,  Klaus},
  year = {2020},
  month = nov,
  pages = {1027–1038}
}

@misc{monai,
Author = {M. Jorge Cardoso and Wenqi Li and Richard Brown and Nic Ma and Eric Kerfoot and Yiheng Wang and Benjamin Murrey and Andriy Myronenko and Can Zhao and Dong Yang and Vishwesh Nath and Yufan He and Ziyue Xu and Ali Hatamizadeh and Andriy Myronenko and Wentao Zhu and Yun Liu and Mingxin Zheng and Yucheng Tang and Isaac Yang and Michael Zephyr and Behrooz Hashemian and Sachidanand Alle and Mohammad Zalbagi Darestani and Charlie Budd and Marc Modat and Tom Vercauteren and Guotai Wang and Yiwen Li and Yipeng Hu and Yunguan Fu and Benjamin Gorman and Hans Johnson and Brad Genereaux and Barbaros S. Erdal and Vikash Gupta and Andres Diaz-Pinto and Andre Dourson and Lena Maier-Hein and Paul F. Jaeger and Michael Baumgartner and Jayashree Kalpathy-Cramer and Mona Flores and Justin Kirby and Lee A. D. Cooper and Holger R. Roth and Daguang Xu and David Bericat and Ralf Floca and S. Kevin Zhou and Haris Shuaib and Keyvan Farahani and Klaus H. Maier-Hein and Stephen Aylward and Prerna Dogra and Sebastien Ourselin and Andrew Feng},
Title = {MONAI: An open-source framework for deep learning in healthcare},
Year = {2022},
Eprint = {arXiv:2211.02701},
}

@article{MalekiVarnosfaderani2024,
  title = {The Role of AI in Hospitals and Clinics: Transforming Healthcare in the 21st Century},
  volume = {11},
  ISSN = {2306-5354},
  url = {http://dx.doi.org/10.3390/bioengineering11040337},
  DOI = {10.3390/bioengineering11040337},
  number = {4},
  journal = {Bioengineering},
  publisher = {MDPI AG},
  author = {Maleki Varnosfaderani,  Shiva and Forouzanfar,  Mohamad},
  year = {2024},
  month = mar,
  pages = {337}
}

@ARTICLE{menze2014multimodal,
  author={Menze, Bjoern H. and Jakab, Andras and Bauer, Stefan and Kalpathy-Cramer, Jayashree and Farahani, Keyvan and Kirby, Justin and Burren, Yuliya and Porz, Nicole and Slotboom, Johannes and Wiest, Roland and Lanczi, Levente and Gerstner, Elizabeth and Weber, Marc-André and Arbel, Tal and Avants, Brian B. and Ayache, Nicholas and Buendia, Patricia and Collins, D. Louis and Cordier, Nicolas and Corso, Jason J. and Criminisi, Antonio and Das, Tilak and Delingette, Hervé and Demiralp, Çağatay and Durst, Christopher R. and Dojat, Michel and Doyle, Senan and Festa, Joana and Forbes, Florence and Geremia, Ezequiel and Glocker, Ben and Golland, Polina and Guo, Xiaotao and Hamamci, Andac and Iftekharuddin, Khan M. and Jena, Raj and John, Nigel M. and Konukoglu, Ender and Lashkari, Danial and Mariz, José António and Meier, Raphael and Pereira, Sérgio and Precup, Doina and Price, Stephen J. and Raviv, Tammy Riklin and Reza, Syed M. S. and Ryan, Michael and Sarikaya, Duygu and Schwartz, Lawrence and Shin, Hoo-Chang and Shotton, Jamie and Silva, Carlos A. and Sousa, Nuno and Subbanna, Nagesh K. and Szekely, Gabor and Taylor, Thomas J. and Thomas, Owen M. and Tustison, Nicholas J. and Unal, Gozde and Vasseur, Flor and Wintermark, Max and Ye, Dong Hye and Zhao, Liang and Zhao, Binsheng and Zikic, Darko and Prastawa, Marcel and Reyes, Mauricio and Van Leemput, Koen},
  journal={IEEE Transactions on Medical Imaging}, 
  title={The Multimodal Brain Tumor Image Segmentation Benchmark (BRATS)}, 
  year={2015},
  volume={34},
  number={10},
  pages={1993-2024},
  doi={10.1109/TMI.2014.2377694}
}

@misc{moawad2024braintumorsegmentationbratsmets,
      title={The Brain Tumor Segmentation (BraTS-METS) Challenge 2023: Brain Metastasis Segmentation on Pre-treatment MRI}, 
      author={Ahmed W. Moawad and Anastasia Janas and Ujjwal Baid and Divya Ramakrishnan and Rachit Saluja and Nader Ashraf and Nazanin Maleki and Leon Jekel and Nikolay Yordanov and Pascal Fehringer and Athanasios Gkampenis and Raisa Amiruddin and Amirreza Manteghinejad and Maruf Adewole and Jake Albrecht and Udunna Anazodo and Sanjay Aneja and Syed Muhammad Anwar and Timothy Bergquist and Veronica Chiang and Verena Chung and Gian Marco Conte and Farouk Dako and James Eddy and Ivan Ezhov and Nastaran Khalili and Keyvan Farahani and Juan Eugenio Iglesias and Zhifan Jiang and Elaine Johanson and Anahita Fathi Kazerooni and Florian Kofler and Kiril Krantchev and Dominic LaBella and Koen Van Leemput and Hongwei Bran Li and Marius George Linguraru and Xinyang Liu and Zeke Meier and Bjoern H Menze and Harrison Moy and Klara Osenberg and Marie Piraud and Zachary Reitman and Russell Takeshi Shinohara and Chunhao Wang and Benedikt Wiestler and Walter Wiggins and Umber Shafique and Klara Willms and Arman Avesta and Khaled Bousabarah and Satrajit Chakrabarty and Nicolo Gennaro and Wolfgang Holler and Manpreet Kaur and Pamela LaMontagne and MingDe Lin and Jan Lost and Daniel S. Marcus and Ryan Maresca and Sarah Merkaj and Gabriel Cassinelli Pedersen and Marc von Reppert and Aristeidis Sotiras and Oleg Teytelboym and Niklas Tillmans and Malte Westerhoff and Ayda Youssef and Devon Godfrey and Scott Floyd and Andreas Rauschecker and Javier Villanueva-Meyer and Irada Pfluger and Jaeyoung Cho and Martin Bendszus and Gianluca Brugnara and Justin Cramer and Gloria J. Guzman Perez-Carillo and Derek R. Johnson and Anthony Kam and Benjamin Yin Ming Kwan and Lillian Lai and Neil U. Lall and Fatima Memon and Mark Krycia and Satya Narayana Patro and Bojan Petrovic and Tiffany Y. So and Gerard Thompson and Lei Wu and E. Brooke Schrickel and Anu Bansal and Frederik Barkhof and Cristina Besada and Sammy Chu and Jason Druzgal and Alexandru Dusoi and Luciano Farage and Fabricio Feltrin and Amy Fong and Steve H. Fung and R. Ian Gray and Ichiro Ikuta and Michael Iv and Alida A. Postma and Amit Mahajan and David Joyner and Chase Krumpelman and Laurent Letourneau-Guillon and Christie M. Lincoln and Mate E. Maros and Elka Miller and Fanny Moron and Esther A. Nimchinsky and Ozkan Ozsarlak and Uresh Patel and Saurabh Rohatgi and Atin Saha and Anousheh Sayah and Eric D. Schwartz and Robert Shih and Mark S. Shiroishi and Juan E. Small and Manoj Tanwar and Jewels Valerie and Brent D. Weinberg and Matthew L. White and Robert Young and Vahe M. Zohrabian and Aynur Azizova and Melanie Maria Theresa Bruseler and Mohanad Ghonim and Mohamed Ghonim and Abdullah Okar and Luca Pasquini and Yasaman Sharifi and Gagandeep Singh and Nico Sollmann and Theodora Soumala and Mahsa Taherzadeh and Philipp Vollmuth and Martha Foltyn-Dumitru and Ajay Malhotra and Aly H. Abayazeed and Francesco Dellepiane and Philipp Lohmann and Victor M. Perez-Garcia and Hesham Elhalawani and Maria Correia de Verdier and Sanaria Al-Rubaiey and Rui Duarte Armindo and Kholod Ashraf and Moamen M. Asla and Mohamed Badawy and Jeroen Bisschop and Nima Broomand Lomer and Jan Bukatz and Jim Chen and Petra Cimflova and Felix Corr and Alexis Crawley and Lisa Deptula and Tasneem Elakhdar and Islam H. Shawali and Shahriar Faghani and Alexandra Frick and Vaibhav Gulati and Muhammad Ammar Haider and Fatima Hierro and Rasmus Holmboe Dahl and Sarah Maria Jacobs and Kuang-chun Jim Hsieh and Sedat G. Kandemirli and Katharina Kersting and Laura Kida and Sofia Kollia and Ioannis Koukoulithras and Xiao Li and Ahmed Abouelatta and Aya Mansour and Ruxandra-Catrinel Maria-Zamfirescu and Marcela Marsiglia and Yohana Sarahi Mateo-Camacho and Mark McArthur and Olivia McDonnell and Maire McHugh and Mana Moassefi and Samah Mostafa Morsi and Alexander Munteanu and Khanak K. Nandolia and Syed Raza Naqvi and Yalda Nikanpour and Mostafa Alnoury and Abdullah Mohamed Aly Nouh and Francesca Pappafava and Markand D. Patel and Samantha Petrucci and Eric Rawie and Scott Raymond and Borna Roohani and Sadeq Sabouhi and Laura M. Sanchez-Garcia and Zoe Shaked and Pokhraj P. Suthar and Talissa Altes and Edvin Isufi and Yaseen Dhemesh and Jaime Gass and Jonathan Thacker and Abdul Rahman Tarabishy and Benjamin Turner and Sebastiano Vacca and George K. Vilanilam and Daniel Warren and David Weiss and Fikadu Worede and Sara Yousry and Wondwossen Lerebo and Alejandro Aristizabal and Alexandros Karargyris and Hasan Kassem and Sarthak Pati and Micah Sheller and Katherine E. Link and Evan Calabrese and Nourel hoda Tahon and Ayman Nada and Yuri S. Velichko and Spyridon Bakas and Jeffrey D. Rudie and Mariam Aboian},
      year={2024},
      eprint={2306.00838},
      archivePrefix={arXiv},
      primaryClass={q-bio.OT},
      url={https://arxiv.org/abs/2306.00838}, 
}

@ARTICLE{BoneMetsReviewe,
AUTHOR={Rich, Joseph M.  and Bhardwaj, Lokesh N.  and Shah, Aman  and Gangal, Krish  and Rapaka, Mohitha S.  and Oberai, Assad A.  and Fields, Brandon K. K.  and Matcuk, George R.  and Duddalwar, Vinay A. },
TITLE={Deep learning image segmentation approaches for malignant bone lesions: a systematic review and meta-analysis},
JOURNAL={Frontiers in Radiology},
VOLUME={3},
YEAR={2023},
URL={https://www.frontiersin.org/journals/radiology/articles/10.3389/fradi.2023.1241651},
DOI={10.3389/fradi.2023.1241651},
ISSN={2673-8740},
}

@Article{MLreviewMets,
AUTHOR = {Jekel, Leon and Brim, Waverly R. and von Reppert, Marc and Staib, Lawrence and Cassinelli Petersen, Gabriel and Merkaj, Sara and Subramanian, Harry and Zeevi, Tal and Payabvash, Seyedmehdi and Bousabarah, Khaled and Lin, MingDe and Cui, Jin and Brackett, Alexandria and Mahajan, Amit and Omuro, Antonio and Johnson, Michele H. and Chiang, Veronica L. and Malhotra, Ajay and Scheffler, Björn and Aboian, Mariam S.},
TITLE = {Machine Learning Applications for Differentiation of Glioma from Brain Metastasis—A Systematic Review},
JOURNAL = {Cancers},
VOLUME = {14},
YEAR = {2022},
NUMBER = {6},
ARTICLE-NUMBER = {1369},
URL = {https://www.mdpi.com/2072-6694/14/6/1369},
PubMedID = {35326526},
ISSN = {2072-6694},
DOI = {10.3390/cancers14061369}
}

\end{document}